\documentclass[runningheads]{llncs}
\usepackage[colorlinks=true, urlcolor=blue, linkcolor=red]{hyperref}
\usepackage{multirow}
\usepackage{color}
\usepackage{cite}
\usepackage{graphicx}
\usepackage{listings}
\usepackage{algorithm}
\usepackage{algpseudocode}
\usepackage{subfigure}
\usepackage{booktabs}
\usepackage{tikz}
\usepackage{makecell}
\usepackage{amssymb}
\usepackage{float}
\def\checkmark{\tikz\fill[scale=0.4](0,.35) -- (.25,0) -- (1,.7) -- (.25,.15) -- cycle;} 
\newcommand{\red}[1]{\textcolor{red}{#1}}
\DeclareUnicodeCharacter{2212}{-}
\begin{document}
\title{Semi-Supervised Semantic Segmentation using Redesigned Self-Training for White Blood Cells}
%
%
\author{Quoc-Vinh Luu\inst{1} \and
Khanh-Duy Le \inst{1} \and
Thanh-Huy Nguyen\inst{2} \and
Thanh-Minh Nguyen\inst{3} \and
Tien-Thinh Nguyen\inst{1}
\and
Quang-Vinh Dinh\inst{4}}
\authorrunning{Quoc-Vinh Luu et al.}
\titlerunning{Semi-Supervised Semantic Segmentation using Redesigned Self-Training}
%
\institute{Ho Chi Minh University of Technology, Vietnam. \and
Taipei Medical University, Taiwan. \and
Ho Chi Minh City Medicine and Pharmacy University, Vietnam. \and
Vietnamese-German University, Vietnam
}
\maketitle              
\begin{abstract}
Artificial Intelligence (AI) in healthcare, especially in white blood cell cancer diagnosis, is hindered by two primary challenges: the lack of large-scale labeled datasets for white blood cell (WBC) segmentation and outdated segmentation methods. These challenges inhibit the development of more accurate and modern techniques to diagnose cancer relating to white blood cells. To address the first challenge, a semi-supervised learning framework should be devised to efficiently capitalize on the scarcity of the dataset available.  In this work, we address this issue by proposing a novel self-training pipeline with the incorporation of FixMatch. Self-training is a technique that utilizes the model trained on labeled data to generate pseudo-labels for the unlabeled data and then re-train on both of them. FixMatch is a consistency-regularization algorithm to enforce the model's robustness against variations in the input image. We discover that by incorporating FixMatch in the self-training pipeline, the performance improves in the majority of cases. Our performance achieved the best performance with the self-training scheme with consistency on DeepLab-V3 architecture and ResNet-50, reaching 90.69\%, 87.37\%, and 76.49\% on Zheng 1, Zheng 2, and LISC datasets, respectively. 

\keywords{Semi-supervised learning  \and Semantic segmentation \and White blood cell}
\end{abstract}
\section{Introduction}
Cancer of the white blood cells (WBC), namely leukemia and lymphoma, are two of the most common types of blood cancers \cite{most-common-blood-cancer}. Certain screening and diagnosis of blood cancers, particularly cancer of the white blood cells (WBC), are still done manually, as in peripheral blood smear analysis. These tests frequently serve the purpose of classifying and counting the number of WBCs. Due to the time-consuming and experts required for labeling, most blood cell datasets lack high-quality annotation. Small-scale public datasets are not sufficient to train robust deep learning to be employed in real-world scenarios.

We intend to make use of a semi-supervised learning technique as we are aware that the field of white blood cell segmentation lacks meaningful datasets. Even with a limited amount of labeled data, we can still construct a robust model by effectively utilizing unlabeled datasets.
In this work, we aim to provide a self-training approach by leveraging unlabeled data to boost the performance of the white blood cell segmentation. Secondly, inspired by FixMatch\cite{fixmatch}, we utilize a mechanism of consistency regularization to aid the self-training framework. Thirdly, we ran extensive experiments to measure the performance of the self-training approach for semi-supervised learning under various settings. 

\section{Related works}
\subsubsection{Medical Image Segmentation.}
In recent years, with the growth of deep learning, medical image analysis has attracted a lot of advancement by automatically diagnosing cancer through radiology\cite{huy1,huy2}, microscopy\cite{cell1}, etc. Noticeably, U-Net\cite{unet} is an autoencoder-based architecture with skip connections to incorporate feature maps from the encoder and decoder. It has been used extensively in segmentation. The combination of U-net with the backbone as Resnet-family\cite{resnet} has shown impressive performance across many medical segmentation tasks and modalities and has become one of the most common baselines for semantic segmentation problems.
\subsubsection{Semi-supervised semantic segmentation} This technique sits between supervised learning, where the whole dataset is labeled, and unsupervised learning, where no data is labeled. Through many years of development, Semi-supervised learning semantic segmentation can be categorized into two main directions: consistency regularization\cite{fixmatch,cct}, and pseudo-labeling\cite{st++}. The self-training-based approach is one of the most promising approaches due to its effective strategy to leverage unlabeled sets. Self-training ST++ \cite{st++} is a sophisticated framework that performs selective re-training by prioritizing reliable images based on holistic prediction-level stability throughout the training course. 


\section{Methodology}

\subsection{Self-training (ST)} 
Similar to Mean teacher, self-training also incorporated the idea of the teacher-student model. But one significant difference is that the teacher and student are trained in two separate stages. The steps for the plain Self-training are as follows:
\begin{enumerate}
    \item Train the teacher model $T$ on labeled data $X_L = \{x_i\}_{i = 1}^L$. 
    \item Use the model that has just been trained on labeled data to generate predictions on the unlabeled data. These predictions in the context of semantic segmentation can be considered as pseudo-masks. This set of pseudo-masks can be formulated as  $\hat{D}_U = \left\{ (x_i, T(x_i)) \right\}_{i=1}^U$
    \item Extend the labeled data ${D}_L = \left\{ (x_i, y_i) \right\}_{i=1}^L$  with the unlabeled ones $\hat{D}_U$ by unioning the labeled set with the unlabeled set, then re-train the student model S on this new set.
\end{enumerate}
In \cite{st++}, the authors propose a stronger baseline for this plain self-training pipeline by incorporating strong data augmentation in the final stage on unlabelled data. The performance upon implementing these is competitive in comparison with other methods at the time.

\begin{figure}[t]
    \centering
    \includegraphics[width = 0.99\linewidth]{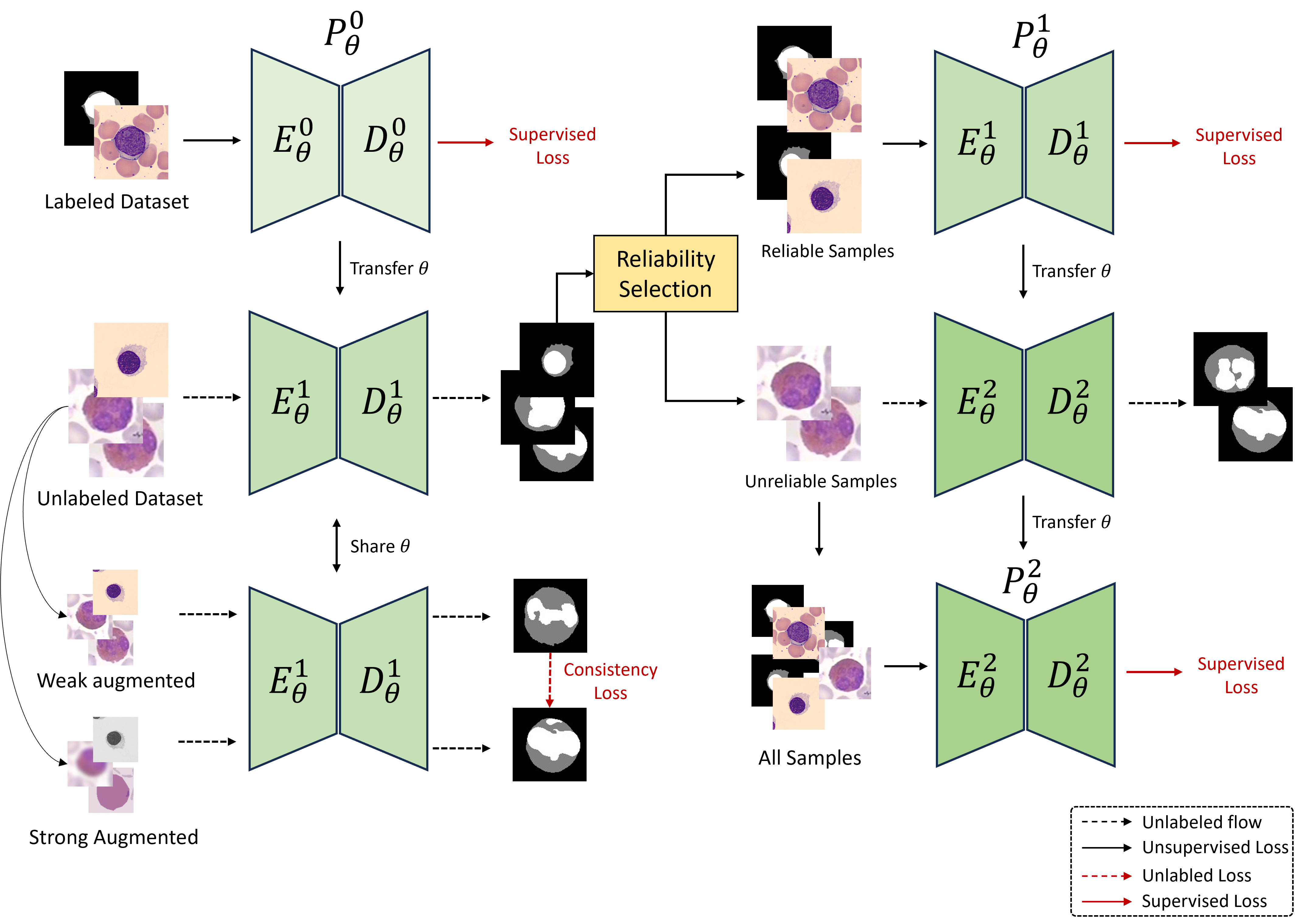}
    \caption{Our proposed semi-supervised semantic segmentation framework. Consistency regularization is incorporated in the first stage of training to boost the capabilities of performing towards unlabeled images with weak-to-strong mechanisms. This incorporation can be done in both the ST and ST++ framework. In this diagram, ST++ is taken as an example}
    \label{figure1}
\end{figure}

\subsection{Self-training with Consistency Regularization}
In \cite{st++}, the authors propose a better self-training scheme, called Self-Training++ (ST++). The innovation in this scheme is that instead of training on all unlabeled images at the same time, the authors divide this stage into two stages by separating unlabeled images into reliable and unreliable ones. Realizing that the ST and ST++ approaches have not yet fully utilized consistency regularization approaches commonly used in semi-supervised learning, we aspire to incorporate FixMatch\cite{fixmatch} in the first supervised training step to train end-to-end framework with both supervised and unsupervised losses. In doing so, we allow the first supervised training stage to take advantage of both labeled and unlabeled data, rather than just labeled data in the original framework. What's more, the weak-to-strong consistency regularization mechanism allows the model to be more robust against strong variations in the unlabeled data.  
\begin{algorithm}[t]
\caption{Supervised training stage in ST and ST++ with FixMatch}
\begin{algorithmic}[1]
    \Require Labeled dataset $D_L = \{(x_i, y_i)\}$, Unlabeled dataset $D_U = \{u_j\}$, Model $M$ parameters $\theta$, Confidence threshold $\tau$, Strong augmentation function $S(\cdot)$, Weak augmentation function $W(\cdot)$
    \State \textbf{Initialize:} Train model $M$ on $L$ for a few epochs

    \For{each training epoch}
        \For{each mini-batch}
            \State Draw a batch of labeled samples $\{(x_i, y_i)\}$ from $D_L$
            \State Apply weak augmentation: $x'_i = W(x_i)$
            \State Draw a batch of unlabeled samples $\{u_j\}$ from $U$
            \State Apply weak augmentation: $u'_j = W(u_j)$
            \State Apply strong augmentation: $u''_j = S(u_j)$
            \State Obtain predictions for weakly-augmented unlabeled data: $p_j = M(u'_j; \theta)$
            \State Generate pseudo-labels: $\hat{y}_j = \arg\max(p_j)$ if $\max(p_j) > \tau$ else ignore

            \State \textbf{Supervised Training Step (incorporating FixMatch):}
            \State Train model on $\{(x'_i, y_i)\}$ using supervised loss
            \State Train model on $\{(u''_j, \hat{y}_j)\}$ using consistency loss
        \EndFor
        \State Update model parameters $\theta$
    \EndFor
    \State \Return Optimized model parameters $\theta$
\end{algorithmic}
\end{algorithm}
\section{Experiments}
\subsection{Dataset}
\textbf{Zheng 1} \cite{dataset1-2} comprises 300 images of white blood cells (WBCs), each measuring 120×120 pixels with a color depth of 24 bits. The images in this dataset predominantly exhibit a yellow background. \textbf{Zheng 2} \cite{dataset1-2} is composed of 100 color images, each with dimensions of 300×300 pixels, acquired from the CellaVision blog. \textbf{The LISC dataset} \cite{lisc} consists of hematological images taken from the peripheral blood of healthy subjects. It was collected from 8 subjects and amounting to 400 samples from 100 slides with a size of 720×576 pixels.
\subsection{Settings}
The experiment was run under two settings: With FixMatch and without FixMatch. Under each of these settings, we employ three frameworks in semi-supervised learning: SupOnly (trained on the labeled dataset only), ST (trained in a standard self-training manner), and ST++(trained in a self-traning++ framework). The labeled set accounts for 1/4 of the size of the training set. This proportion is in accordance with the setting used in the semi-supervised learning setting, such as ST++ \cite{st++}.

Without consistency, the batch size for datasets Zheng 1, Zheng 2, and LISC is 16. With the consistency incorporated, the batch size for the labeled and the unlabeled data of Zheng 1 is 16. For Zheng 2 and LISC, the labeled data's batch size is 16 and the unlabeled data's batch size is 8. The network architectures are DeepLabV3+, DeepLabv2, and PSPNet. The choice of the DeepLab networks was because of their atrous convolution which enables wider context absorption. PSPNet also aggregates context at different scales. All of these attributes enable the networks to segment the nucleus and cytoplasm of the cells effectively. Each network architecture consists of the ResNet-50 and ResNet-101 backbone, but DeepLabv2 only has the ResNet-101 backbone. The learning rate for three datasets is 0.009 and follows a poly decay schedule $lr = baselr \times \left(1 - \frac{iter}{total\_iter}\right)^{0.9}.$\\
\section{Results}
\begin{table}[t]
\caption{The mIOU of the models run under different networks with different backbones.}
\label{result-table}
\resizebox{\textwidth}{!}{%
\begin{tabular}{|c|c|c|ccccccccc|}
\hline
Model                       & Backbone                     & Mode                                & \multicolumn{9}{c|}{Dataset}                                                                                                                                                                                                                                                                                                                                                                        \\ \hline
                            &                              & \multicolumn{1}{l|}{}                   & \multicolumn{3}{c|}{Zheng   1}                                                                                                & \multicolumn{3}{c|}{Zheng 2}                                                                                                                          & \multicolumn{3}{c|}{LISC}                                                                                \\ \cline{4-12} 
                            &                              & \multicolumn{1}{l|}{\multirow{-2}{*}{}} & \multicolumn{1}{c|}{SupOnly} & \multicolumn{1}{c|}{ST}             & \multicolumn{1}{c|}{ST++}                                  & \multicolumn{1}{c|}{SupOnly} & \multicolumn{1}{c|}{ST}                                    & \multicolumn{1}{c|}{ST++}                                  & \multicolumn{1}{c|}{SupOnly} & \multicolumn{1}{c|}{ST}             & ST++                                \\ \cline{3-12} 
                            &                              & without FixMatch                     & \multicolumn{1}{c|}{89.49}   & \multicolumn{1}{l|}{90.71}          & \multicolumn{1}{l|}{90.02}                                 & \multicolumn{1}{c|}{86.19}   & \multicolumn{1}{l|}{86.43}                                 & \multicolumn{1}{l|}{87.34}                                 & \multicolumn{1}{c|}{75.2}    & \multicolumn{1}{l|}{76.67}          & \multicolumn{1}{l|}{77.64}          \\ \cline{3-12} 
\multirow{-4}{*}{DeepLabv3} & \multirow{-4}{*}{ResNet-50}  & with FixMatch                        & \multicolumn{1}{c|}{87.81}   & \multicolumn{1}{l|}{90.04}          & \multicolumn{1}{l|}{{\color[HTML]{333333} \red{90.69}}} & \multicolumn{1}{c|}{84.19}   & \multicolumn{1}{l|}{{\color[HTML]{333333} \red{87.02}}} & \multicolumn{1}{l|}{{\color[HTML]{333333} \red{87.37}}} & \multicolumn{1}{c|}{73.96}   & \multicolumn{1}{l|}{76.48} & \multicolumn{1}{l|}{76.49} \\ \hline
                            &                              & without FixMatch                     & \multicolumn{1}{c|}{88.86}   & \multicolumn{1}{c|}{89.83}          & \multicolumn{1}{c|}{89.99}                                 & \multicolumn{1}{c|}{86.03}   & \multicolumn{1}{c|}{86.35}                                 & \multicolumn{1}{c|}{87.22}                                 & \multicolumn{1}{c|}{75.95}   & \multicolumn{1}{c|}{75.87}          & 76.55                               \\ \cline{3-12} 
\multirow{-2}{*}{DeepLabv3} & \multirow{-2}{*}{ResNet-101} & with FixMatch                        & \multicolumn{1}{c|}{86.61}   & \multicolumn{1}{c|}{89.8}           & \multicolumn{1}{c|}{\red{90.29}}                        & \multicolumn{1}{c|}{85.15}   & \multicolumn{1}{c|}{\red{87.28}}                        & \multicolumn{1}{c|}{87.09}                                 & \multicolumn{1}{c|}{75.09}   & \multicolumn{1}{c|}{\red{77.02}} & \red{76.6}                       \\ \hline
                            &                              & without FixMatch                   & \multicolumn{1}{c|}{82.54}   & \multicolumn{1}{c|}{84.21}          & \multicolumn{1}{c|}{85.01}                                 & \multicolumn{1}{c|}{74.46}   & \multicolumn{1}{c|}{79.7}                                  & \multicolumn{1}{c|}{78.74}                                 & \multicolumn{1}{c|}{72.2}    & \multicolumn{1}{c|}{75.27}          & 74.84                               \\ \cline{3-12} 
\multirow{-2}{*}{DeepLabv2} & \multirow{-2}{*}{ResNet-101} & with FixMatch                        & \multicolumn{1}{c|}{80.08}   & \multicolumn{1}{c|}{\red{85.05}} & \multicolumn{1}{c|}{\red{85.07}}                        & \multicolumn{1}{c|}{71.47}   & \multicolumn{1}{c|}{79.32}                                 & \multicolumn{1}{c|}{\red{80.38}}                        & \multicolumn{1}{c|}{70.47}   & \multicolumn{1}{c|}{74.71}          & \red{75.75}                      \\ \hline
                            &                              & without FixMatch                     & \multicolumn{1}{c|}{80.27}   & \multicolumn{1}{c|}{83.08}          & \multicolumn{1}{c|}{83.64}                                 & \multicolumn{1}{c|}{75.76}   & \multicolumn{1}{c|}{80.2}                                  & \multicolumn{1}{c|}{80.92}                                 & \multicolumn{1}{c|}{71.17}   & \multicolumn{1}{c|}{74.69}          & 75.12                               \\ \cline{3-12} 
\multirow{-2}{*}{PSPNet}    & \multirow{-2}{*}{ResNet-50}  & with FixMatch                        & \multicolumn{1}{c|}{79.02}   & \multicolumn{1}{c|}{\red{83.44}} & \multicolumn{1}{c|}{\red{83.76}}                        & \multicolumn{1}{c|}{74.95}   & \multicolumn{1}{c|}{\red{80.31}}                        & \multicolumn{1}{c|}{\red{81.79}}                        & \multicolumn{1}{c|}{71.53}   & \multicolumn{1}{c|}{\red{75.5}}  & \red{75.39}                      \\ \hline
                            &                              & without FixMatch                     & \multicolumn{1}{c|}{81.56}   & \multicolumn{1}{c|}{83.93}          & \multicolumn{1}{c|}{84.3}                                  & \multicolumn{1}{c|}{77.76}   & \multicolumn{1}{c|}{80.39}                                 & \multicolumn{1}{c|}{81.35}                                 & \multicolumn{1}{c|}{72.91}   & \multicolumn{1}{c|}{75.19}          & 75.17                               \\ \cline{3-12} 
\multirow{-2}{*}{PSPNet}    & \multirow{-2}{*}{ResNet-101} & with FixMatch                        & \multicolumn{1}{c|}{80.69}   & \multicolumn{1}{c|}{\red{84.56}} & \multicolumn{1}{c|}{\red{84.58}}                        & \multicolumn{1}{c|}{77.33}   & \multicolumn{1}{c|}{79.8}                                  & \multicolumn{1}{c|}{80.11}                                 & \multicolumn{1}{c|}{71.35}   & \multicolumn{1}{c|}{74.96}          & 74.82                               \\ \hline
\end{tabular}%
}
\end{table}
\subsection{Performance across SupOnly, ST, ST++ with and without FixMatch incorporated} 
As described in Table \ref{result-table}, there is a performance improvement when transitioning from Supervised-Only (SupOnly) to Self-Training (ST) and further to ST++, with ST++ demonstrating the highest level of performance. This is not surprising, since each framework is an improvement upon one another. With SupOnly, the model only takes advantage of labeled data, while with ST and ST++, the model also harnesses the power of unlabeled data. Furthermore, ST++ has more discriminative power over ST, since it separates the unlabeled data into reliable and unreliable ones. 
\begin{table}
    \caption{Noticeable decrease when incorporating FixMatch into the model. The left and right column denotes the absence and presence of FixMatch respectively}
    \label{table2}
    \centering
    \begin{tabular}{|c|c|c|c|c|c|c|} \hline 
 Model + Backbone& \multicolumn{2}{|c|}{Zheng 1}& \multicolumn{2}{|c|}{Zheng 2}& \multicolumn{2}{|c|}{LISC}\\ \hline 
         &  $\times$&  $\checkmark$&  $\times$&  $\checkmark$& $\times$&$\checkmark$\\ \hline 
         DeepLabv3+ - ResNet50&  89.49&  87.81&  86.19&  84.19&  75.2&73.96\\ \hline 
 DeepLabv3+ - ResNet101& 88.86& 86.61& 86.03& 85.15& 75.95&75.09\\ \hline 
 DeepLabv2 - ResNet101& 82.54& 80.08& 74.46& 71.47& 72.2&70.47\\ \hline 
 PSPNet - ResNet50& 80.27& 79.02& 75.76& 74.95& 71.17&71.53\\ \hline 
 PSPNet - ResNet101& 81.56& 80.69& 77.76& 77.33& 72.91&71.35\\ 
 \hline
    \end{tabular}
    \label{tab:my_label}
\end{table}
With FixMatch incorporated, the performance of the ST and ST++ framework improves in the majority of cases, indicating that the incorporation of FixMatch is beneficial in the standard ST and ST++ framework. FixMatch training alongside the supervised training stage helps the model get a glimpse into the unlabeled data before actually training on them. In ST, the performance does not improve in as many cases as it does with ST++, primarily because it does not have separate stages to classify unlabeled images into reliable and unreliable ones. These stages in ST++ benefit tremendously from the FixMatch inclusion in the first stage, thus helping the model gain improvement in performance.

\subsection{Performance on the supervised training stage when incorporating FixMatch}

When incorporating FixMatch into the supervised training stage, there was a noticeable decrease in performance on the labeled images, as observed in Table \ref{table2}. This is a commonly seen trade-off when we try to do two tasks in 1 stage. The performance dropped on supervised images is necessary for the model to perform well on unlabeled ones. 

However, the performance in the later stage remains the same or even improves, suggesting that the trade-off in the first stage is necessary to improve the performance when extending the labeled images with unlabeled ones.
\begin{figure}[h]
    \centering
    \vspace{-3mm}
    \includegraphics[width = 0.83\linewidth]{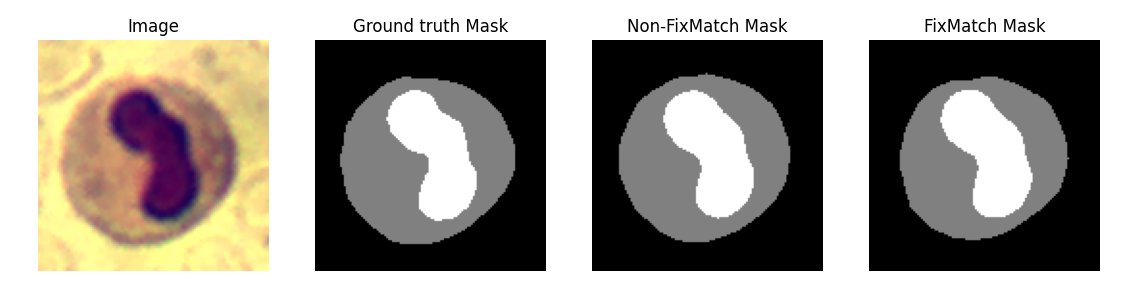}
    \caption{Sample of Ground truth mask, Non-FixMatch, and FixMatch mask generated from the dataset Zheng 1.}
    \label{figure1}
    \vspace{-5mm}
\end{figure}
\subsection{The quality of the pseudo-masks produced when incorporating FixMatch}
\begin{figure}[h]
    \centering
    \vspace{-3mm}
    \includegraphics[width = 0.83\linewidth]{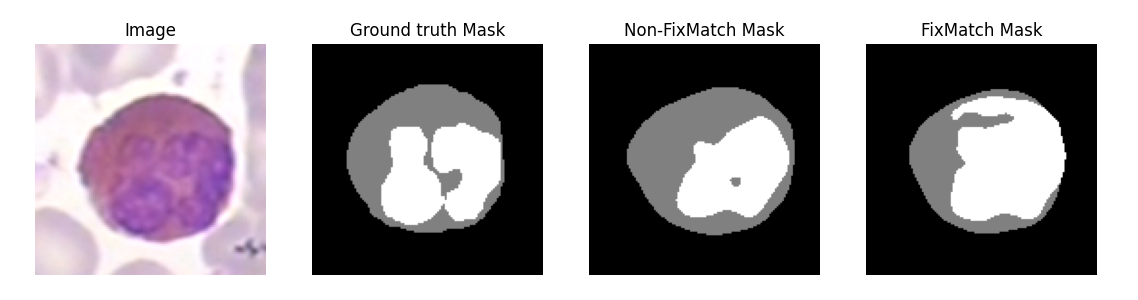}
    \caption{Deviations from the Ground truth mask of Non-FixMatch and FixMatch mask generated from the LISC dataset.}
    \label{figure2}
\end{figure}

The quality of the pseudo-masks between the two settings of FixMatch and non-FixMatch approaches varies. On the one hand, the pseudo-masks of easy, reliable images remain essentially the same between the two settings and on par with that of the ground truth mask, as illustrated in Figure \ref{figure1}.
However, for more complicated, unreliable images, the masks generated can have a large deviation from the ground truth mask, as can be seen in Figure \ref{figure2}.

This suggests that incorporating FixMatch into the model has yet to be able to take on the nuances presented in challenging images. Therefore, there should be a more domain-specific way to account for these nuances.
\subsection{Generalization capability of the proposed approach}
\begin{figure}[h]
    \centering
    \includegraphics[width=0.83\linewidth]{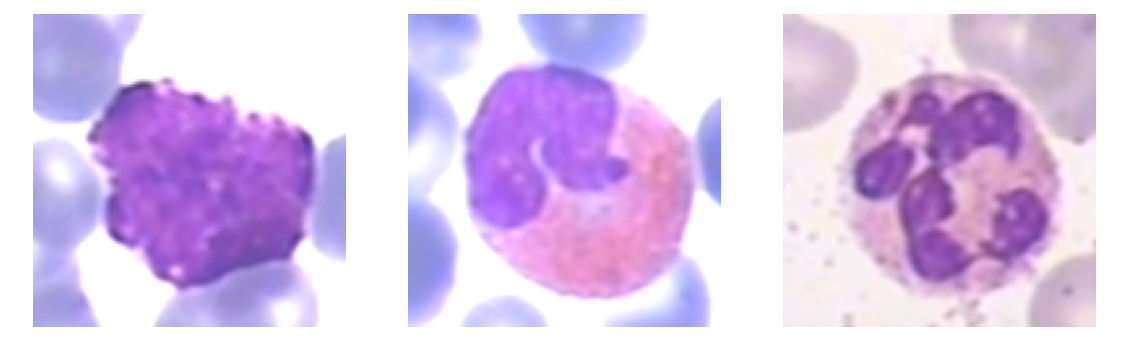}
    \caption{LISC's variability between its images leads to the model's significantly reduced performances}
    \label{lisc_three_images}
\end{figure}

The proposed approach can be used in any dataset since the framework only includes general approaches commonly used in semi-supervised settings, namely self-training and FixMatch. However, the performance metrics are higher for datasets that are less complex compared to those that are more challenging. The performance of the proposed approach on Zheng 1 \cite{dataset1-2} and Zheng 2 \cite{dataset1-2} is impressive, while the performance on LISC \cite{lisc} is low. This could be explained by LISC's huge variability between its images \ref{lisc_three_images}. Therefore, the proposed approach has yet to resolve the intra-variation in challenging datasets and achieve consistent performance across different datasets. 
\section{Conclusion}
\vspace{-1mm}
In this paper, we propose a novel method for incorporating consistency regularization into standard and enhanced self-training approaches. In most cases, we find that this incorporation is beneficial. We also discover that there is a trade-off between excelling solely on labeled images and excelling on labeled images while using consistency regularization on unlabeled images. Furthermore, even with FixMatch included, the ST and ST++ approaches have yet to address the nuances of difficult images. To address these nuances, we plan to develop more domain-specific approaches in white blood cell segmentation in the future.

\begin{credits}

%
%
%
%

\end{credits}
\end{document}